%% file: main.tex
\documentclass[runningheads]{llncs}

\usepackage{tikz}
\usepackage{comment}
\usepackage{amsmath,amssymb}
\usepackage{textcomp}
\usepackage{gensymb}
\usepackage{color}

\usepackage[accsupp]{axessibility}

\usepackage{epsfig}
\usepackage{graphicx}
\usepackage{wrapfig}
\usepackage{subfigure}

\usepackage{multirow}
\usepackage{rotating}
\usepackage{booktabs}
\usepackage{tabularx}
\usepackage{colortbl}

\usepackage{hyperref}
\hypersetup{breaklinks=true,colorlinks}

\usepackage{xspace}
\usepackage[T1]{fontenc}
\usepackage{paralist}

\input{macros.tex}

\begin{document}
\pagestyle{headings}
\mainmatter

\title{Sim-2-Sim Transfer for Vision-and-Language Navigation in Continuous Environments}
\titlerunning{Sim-2-Sim Transfer for VLN-CE}
\author{Jacob Krantz \and Stefan Lee}
\authorrunning{J. Krantz and S. Lee}
\institute{
    Oregon State University
    \newline\email{\{krantzja,leestef\}@oregonstate.edu}
}
\maketitle

\input{sections/01_abstract}
\input{sections/02_intro}
\input{sections/03_related}
\input{sections/04_task}
\input{sections/05_method}
\input{sections/06_results}
\input{sections/07_conclusion}

\input{sections/08_acknowledgement}

\clearpage
\bibliographystyle{splncs04}
\bibliography{bib/strings,bib/main}

\clearpage
\input{sections/09_supp}

\end{document}

%% file: macros.tex
\newcolumntype{s}{>{\columncolor[gray]{.85}[.5\tabcolsep]}c}

\newcommand{\csection}[1]{
    \section{#1}
}

\newcommand{\csubsection}[1]{
    \subsection{#1}
}

\newcommand{\xhdr}[1]{\vspace{2pt}\noindent\textbf{#1}}

\newcommand{\tabref}[1]{Tab.~\ref{#1}\xspace}
\newcommand{\figref}[1]{Fig.~\ref{#1}\xspace}
\newcommand{\secref}[1]{Sec.~\ref{#1}\xspace}

\newcommand{\etal}{et al.\xspace}

\newcommand{\ie}{i.e.\xspace}
\newcommand{\vs}{vs.\xspace}

\def\CircleArrowright{\ensuremath{
  \rotatebox[origin=c]{310}{$\circlearrowright$}}}
\newcommand{\vlnbert}{VLN$\protect\CircleArrowright$BERT}

\newcommand{\valtrain}{Train$^{*}$}

\newcommand{\vlncebert}{VLN-CE$\protect\CircleArrowright$BERT}

\newcommand{\tightcaption}{\vspace{-1.5\baselineskip}}

%% file: sections/01_abstract.tex
\begin{abstract}
Recent work in Vision-and-Language Navigation (VLN) has presented two environmental paradigms with differing realism -- the standard VLN setting built on topological environments where navigation is abstracted away \cite{anderson2018vision}, and the VLN-CE setting where agents must navigate continuous 3D environments using low-level actions \cite{krantz2020beyond}. Despite sharing the high-level task and even the underlying instruction-path data, performance on VLN-CE lags behind VLN significantly. In this work, we explore this gap by transferring an agent from the abstract environment of VLN to the continuous environment of VLN-CE. We find that this sim-2-sim transfer is highly effective, improving over the prior state of the art in VLN-CE by +12\% success rate. While this demonstrates the potential for this direction, the transfer does not fully retain the original performance of the agent in the abstract setting. We present a sequence of experiments to identify what differences result in performance degradation, providing clear directions for further improvement.
\keywords{vision-and-language navigation (VLN), embodied AI}
\end{abstract}

%% file: sections/02_intro.tex
\csection{Introduction}

Vision-and-Language Navigation (VLN) is a popular instruction-guided navigation task where a vision-equipped agent must navigate to a goal location in an never-before-seen environment by following a path described by a natural language instruction. In the standard VLN setting, the environment is abstracted as a topology of interconnected panoramic images (called a nav-graph) such that navigation amounts to traversing the graph by iteratively selecting among a small set of neighboring node locations at each time step. In effect, this induces a prior of possible locations for the agent and assumes perfect navigation between nodes. Recent work has identified that these assumptions do not reflect the challenges a deployed system would experience in a real environment. To reduce this gap, Krantz \etal \cite{krantz2020beyond} introduced the VLN in Continuous Environments (VLN-CE) setting which instantiates VLN in a 3D simulator and drops the nav-graph assumptions -- requiring agents to navigate with low-level actions. 

So far, VLN-CE has proven to be significantly more challenging than its more abstract counterpart, with published work reporting episode success rates less than half of those reported in standard VLN. Typically, models for VLN-CE have been end-to-end systems trained to directly predict low-level actions (or nearby waypoints) from language and observations. As such, the lower performance compared to VLN is commonly ascribed to the challenge of learning navigation and language grounding jointly in a long-horizon task. This suggests two avenues towards improvement -- developing more capable models directly in VLN-CE or exploring transfer of knowledge from VLN to VLN-CE. Given the significant differences between the abstract and continuous settings, the potential of sim-2-sim transfer has so far been under-explored and uncertain.  

In this work, we explore the sim-2-sim transfer of a VLN agent to VLN-CE. To support this transfer, we build a modular harness such that any VLN agent can be run in VLN-CE. Analogous to sim-2-real experiments from \cite{anderson2020sim}, this harness includes a local navigation policy and a subgoal generation module to mimic nav-graph provided candidates. We present a sequence of systematic experiments to quantify what differences between the two task settings are most impactful. Our final model demonstrates that transfer can lead to significant improvements over end-to-end VLN-CE-trained methods -- achieving an increase of +12\% success rate over prior published state-of-the-art (32\% vs 44\%) on the VLN-CE test set.

Along the way, our analysis quantifies the effect of (1) differences between VLN and VLN-CE data subsets, (2) the visual domain gap between real panoramas and simulated renders in VLN-CE, (3) error induced by navigation policies, and (4) subgoal candidate generation -- identifying fruitful directions for future work. Through an analysis of our model's errors, we identify that episodes containing significant elevation change (\ie stairs) are a challenge for our model.

Overall, our results demonstrate an alternative research direction for improving instruction-guided navigation agents in continuous environments. We show transfer with compelling results and present the community with clear remaining challenges. We will release our model-agnostic VLN-2-VLNCE harness code to facilitate research in this area -- helping to unify progress between both settings.\\

\noindent\textbf{Contributions.} To summarize the contributions of this work:\\[-5pt]
\begin{compactitem}[\hspace{5pt}--]
\item We present a first-of-its-kind demonstration that instruction-following agents trained in the abstract VLN setting can effectively transfer to VLN-CE ---\\ setting a new state of the art in VLN-CE by +12\% success rate.\\[-5pt]  
\item Through systematic experiments, we quantify performance loss for key steps of the transfer process -- providing insight to the community about existing gaps in our transfer paradigm that could be improved.\\[-5pt]
\item We develop a modular VLN-2-VLNCE transfer harness that will be released to the community
to spur innovation on transfer techniques and help share progress from VLN to VLN-CE.
\end{compactitem}

%% file: sections/03_related.tex
\csection{Related Work}

\xhdr{VLN Agents.}
The VLN task  \cite{anderson2018vision} has received significant attention in recent years with the continual improvement of benchmark models. A higher-level panoramic action space was introduced by Fried \etal \cite{fried2018speaker} and widely adopted in follow-up works. Early efforts proposed cross-modal grounding of vision and language  \cite{wang2019reinforced}. Pre-training and data augmentation methods can mitigate issues regarding limited data \cite{fried2018speaker,tan2019learning} and challenges of jointly processing vision, language, and action \cite{hao2020towards}. Recently, transformer-based architectures and multi-step training routines have demonstrated superior performance \cite{majumdar2020improving,hong2021vln,chen2021history}. \vlnbert\ \cite{hong2021vln} is one such model that introduced a recurrent history context to a transformer-based agent. We adopt \vlnbert\ as the core VLN agent in this work. With continual interest in improving topological instruction followers, we establish expectations and infrastructure for their transfer to continuous environments.

\xhdr{VLN-CE Agents.}
Recognizing the unrealistic affordances of navigation graphs, Krantz \etal \cite{krantz2020beyond} proposed replacing the topological definition of VLN with lower-level actions in continuous environments. Initial end-to-end baselines that directly predict lower-level actions were modeled after sequence-to-sequence \cite{anderson2018vision} and cross-modal \cite{wang2019reinforced} VLN methods and demonstrated significantly lower performance than in VLN. Raychaudhuri \etal \cite{raychaudhuri2021language} proposed language-aligned path supervision to better guide off-the-path decision-making. Krantz \etal \cite{krantz2021waypoint} explored a higher-level waypoint action space trained with deep reinforcement learning, finding increased performance but still far below VLN. Other works exploited continuous environments to study topological map generation \cite{chen2021topological} and semantic mapping for navigation \cite{irshad2021sasra}. Rather than train agents from scratch in VLN-CE, we analyze the transfer of pre-trained VLN agents to continuous environments.

\xhdr{Sim-2-Real Transfer of Instruction Following.} 
Transferring skills learned in simulation to reality is a critical step for embodied agents \cite{kadian_sim2realgap_2020,gordon2019splitnet,deitke2020robothor,blukis2020learning}. The most similar to our work is that of Anderson \etal \cite{anderson2020sim}, which directly transferred VLN agents to reality. We analyze the intermediate step: sim-2-sim transfer from high-level simulation to more realistic, lower-level simulation. The study of Anderson \etal \cite{anderson2020sim} emulated the VLN action space with subgoal candidate prediction and performed navigation with the Robot Operating System (ROS)\cite{ros}. They found a more than 2x drop in performance when evaluating agents in reality, showing that effective VLN sim-2-real transfer remains an open research question. We identify similar challenges in sim-2-sim transfer as Anderson \etal \cite{anderson2020sim} did in sim-2-real. We diagnose and begin mitigating these issues in VLN-CE using methods infeasible in VLN and impractical in reality.

%% file: sections/04_task.tex
\csection{Task Descriptions}

In this section, we describe both the vision-and-language navigation task (VLN) and its continuous-environment counterpart, VLN-CE.

\xhdr{Vision-and-Language Navigation (VLN).} VLN was proposed by Anderson, \etal \cite{anderson2018vision}, in which agents navigate previously-unseen indoor environments. The task is to follow a path described in natural language and stop at the goal. Instructions are unstructured and crowd-sourced in English to form the Room-to-Room (R2R) dataset. VLN takes place in a topological simulator, where a pre-defined nav-graph dictates allowable navigation through edge connections.

At each time step $t$, a VLN agent receives the  instruction $\mathcal{I}$, a panoramic RGB observation $\mathcal{O}_t$, and a set of $n$ subgoal candidates $\mathcal{C}_{t}^{(1...n)}$. The instruction $\mathcal{I}$ is a sequence of $L$ words $(w_0, w_1, \dots, w_L)$. The panoramic observation $\mathcal{O}_t$ is represented as 36 views such that 12 equally-spaced horizontal headings (0$\degree$, 30$\degree$, \dots, 360$\degree$) are represented at each of 3 elevation angles (-30$\degree$, 0$\degree$, +30$\degree$). Each image frame in $\mathcal{O}_t$ has a resolution of 480x640 with a horizontal field of view of 75$\degree$. Each subgoal candidate $\mathcal{C}_{t}^{i}$ is represented by a relative heading angle $\theta_{t}^{i}$, an elevation angle $\phi_{t}^{i}$, and the observation view frame $\mathcal{O}_t^{j}$ that most closely centers the candidate: $\mathcal{C}_{t}^{i} = (\theta_{t}^{i}, \phi_{t}^{i}, \mathcal{O}_t^{j})$. The VLN agent then navigates by selecting a candidate. The environment then transitions and produces observations for time $t+1$. This repeats until the agent issues the \texttt{STOP} action.

\xhdr{VLN in Continuous Environments (VLN-CE).}
VLN-CE as proposed by Krantz \etal \cite{krantz2020beyond} converts the topologically-defined VLN task into a continuous-environment task more representative of real-world navigation. Instead of selecting and navigating by environment-provided subgoal candidates, agents in VLN-CE must navigate a continuous-valued 3D mesh by selecting actions from a lower-level action space (\texttt{FORWARD} 0.25m, \texttt{TURN-LEFT} 15$\degree$, \texttt{TURN-RIGHT} 15$\degree$, \texttt{STOP}). In this work, we seek to port VLN agents into continuous environments. Thus, we adopt what we can of the VLN observation space: the instruction $\mathcal{I}$ and the panoramic observation $\mathcal{O}_t$. We introduce a 2D laser scan which we describe in \secref{sec:sgm}. This provides our subgoal generation module with 2D occupancy. Such a sensor is commonly used for both localization and mapping in real-world navigation stacks \cite{ros}. In our analysis below, we perform intermediate experiments that may allow oracle navigation or topological subgoal candidates but note that these experiments do not result in admissible VLN-CE agents.

\xhdr{Room-to-Room Dataset.} The Room-to-Room dataset (R2R) consists of $7,189$ shortest-path trajectories across all train, validation, and test splits \cite{anderson2018vision}. The VLN-CE dataset is a subset of R2R consisting of $5,611$ trajectories traversible in continuous environments (78\% of R2R) \cite{krantz2020beyond}. For both tasks, we report performance on several validation splits. Val-Seen contains episodes with novel paths and instructions but from scenes observed in training. Val-Unseen contains novel paths, instructions, and scenes. \valtrain\ is a random subset of the training split derived by Hong \etal \cite{hong2021vln} to support further analysis.

\xhdr{Matterport3D Scene Dataset.} Both VLN and VLN-CE use the Matterport3D (MP3D) Dataset \cite{Matterport3D} which consists of 90 scenes, 10,800 panoramic RGBD images, and 90 3D mesh reconstructions. VLN agents interact with MP3D through a connectivity graph that queries panoramic images. VLN-CE agents interact with the 3D mesh reconstructions through the Habitat Simulator \cite{habitat19arxiv}.

\xhdr{Evaluation Metrics.} We report evaluation metrics standard to the public leaderboards of both VLN and VLN-CE. These include trajectory length (\texttt{TL}), navigation error (\texttt{NE}), oracle success rate (\texttt{OS}), success rate (\texttt{SR}), and success weighted by inverse path length (\texttt{SPL}) as described in \cite{anderson2018evaluation,anderson2018vision}. We consider \texttt{SR} the primary metric for comparison and \texttt{SPL} for evaluating path efficiency. Both \texttt{SR} and \texttt{SPL} determine success using a $3.0$m distance threshold between the agent's terminal location and the goal. We report the same metrics across both VLN and VLN-CE to enable empirical diagnosis of performance differences. We note that distance is computed differently between VLN and VLN-CE resulting in subtle differences in evaluation for the same effective path. In VLN, the distance between nodes is Euclidean\footnote{As defined by the \href{https://github.com/peteanderson80/Matterport3DSimulator}{Matterport3D Simulator} used in VLN.}, whereas in VLN-CE, distances are geodesic as computed on the 3D navigation mesh. This tends to result in longer path lengths in VLN-CE and a drop in \texttt{SPL}.

%% file: sections/05_method.tex
\csection{Porting a VLN agent to Continuous Environments}

\begin{figure}[t]
    \centering
    \includegraphics[width=\textwidth]{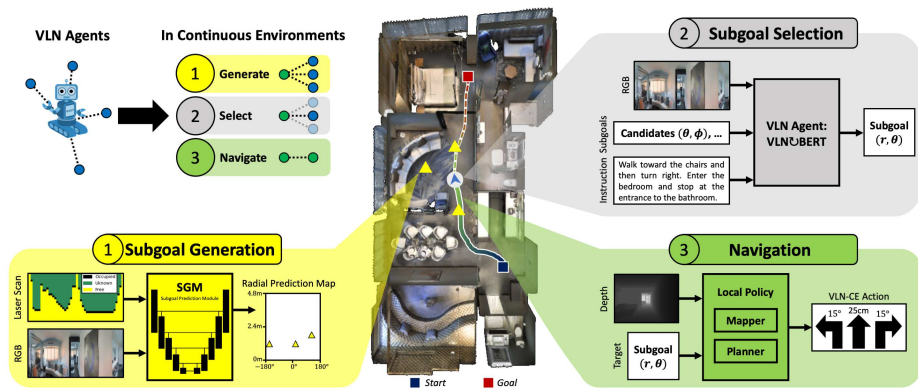}
    \caption{We diagnose and quantify the VLN to VLN-CE gap by operating a topological VLN agent in continuous environments. We emulate the VLN observation space by generating subgoal candidates from egocentric observation (\secref{sec:sgm}) and emulate the action space by calling a map-based navigation policy (\secref{sec:nav_agents}).}
    \label{fig:harness}
\end{figure}

To perform a sim-2-sim transfer, we emulate the VLN action space in VLN-CE by developing a harness consisting of a lower-level navigation policy and a subgoal generation module. An overview can be seen in \figref{fig:harness}. Our harness follows the structure used in sim-2-real experiments by Anderson \etal \cite{anderson2020sim}. We develop our harness such that subgoal generation modules, VLN agents, and navigation policies are all modular components for drop-in replacement and evaluation. This not only enables the analysis in this work, but the released codebase will ease sim-2-sim transfer of future VLN agents to continuous environments. 

\csubsection{Core VLN Agent}
We adopt \vlnbert\ \cite{hong2021vln}, a VLN agent that performs near the state-of-the-art\footnote{\href{https://eval.ai/web/challenges/challenge-page/97/overview}{eval.ai/web/challenges/challenge-page/97}} on R2R Test. This agent uses observation and action spaces typical of recent work in VLN. \vlnbert\ has a recurrence-modified transformer architecture and encodes RGB vision with a ResNet-152 trained on Places365 \cite{he_cvpr16,zhou2017places}. We use \vlnbert\ as our core VLN agent for all experiments but note that our harness and related design choices are agnostic to the choice of VLN agent.

\csubsection{Navigation Policies}
\label{sec:nav_agents}

With the abstracted action space in VLN, navigation between subgoal locations is effectively performed by teleportation. In contrast, continuous settings require a navigation policy to convey the agent between subgoals. This navigation sub-task can be considered a short-range version of PointGoal navigation \cite{anderson2018evaluation} given that the average distance between nodes in the VLN nav-graph is 2.25m. To evaluate the impact of navigation between subgoals in VLN-CE, we first establish upper bounds with two oracle methods: teleportation and an oracle policy. We then evaluate with a local policy admissible in VLN-CE. We describe these navigation policies as follows.

\xhdr{Teleportation.} A teleportation action involves translation to the target location and rotation to face away from the previous location. In our teleportation policy, the heading is snapped to the nearest global 30$\degree$ increment to match the heading discretization in VLN. This policy matches the navigation assumption made in VLN and serves as both an upper bound on navigation performance and a confirmation that the VLN agent is operating as expected in VLN-CE.

\xhdr{Oracle Policy.} An oracle navigation policy has access to the navigation mesh used by the simulator to take optimal actions from the VLN-CE action space. The input to our oracle policy is 3D coordinates that exist on the navigation mesh. Analytical search using the A* algorithm then determines an action plan. We set a stopping distance threshold of 0.15m to the goal which we empirically determine results in minimal navigation error and minimal action jitter near the goal. The Oracle Policy serves as an upper bound on navigation performance to policies operating under the same action space.

\xhdr{Local Policy.} We employ a mapping and planning navigation policy based on \cite{chaplot2020learning} to convey the agent to selected subgoal locations as shown in \figref{fig:harness} ``Navigation''. The Local Policy receives a target location specified by a distance and relative heading $(r, \theta)$. At each time step, a local occupancy map is aggregated from the geometric projection of an egocentric depth observation. The unexplored map area is assumed to be free space. We then plan a path to the target location using the Fast Marching Method (FMM) which produces position coordinates along the shortest path lying 0.25m away from the agent. Following \cite{hahn2021no}, we greedily decode a discrete action to approach this point. We adopt the VLN-CE action space and stop once the agent is within 0.15m of the target or 40 actions have been taken. The occupancy map is re-initialized for each new target.

\csubsection{Subgoal Candidate Generation}
\label{sec:sgm}

The VLN action space requires a set of subgoal candidates to select from. Where VLN uses neighboring nodes in the nav-graph, such oracle information does not exist in VLN-CE. As shown in \figref{fig:harness} ``Subgoal Generation'', we emulate the presence of a nav-graph by having a learned module predict subgoal candidates from observations. Anderson \etal \cite{anderson2020sim} developed a subgoal generation module (SGM) for VLN sim-2-real transfer. We adopt SGM and modify it for VLN-CE.

The observation space of the SGM is panoramic RGB vision (same as the VLN agent), supplemented with a 2D laser scanner for range-finding. We emulate a 360$\degree$ laser scanner in the Habitat Simulator mounted at 0.24m above ground level. We note that such scanners are commercially available. As shown in \figref{fig:harness} ``Laser Scan'', the laser scan is represented as a radial obstacle map limited to a $4.8$m range. The obstacle map size is 24x48 with 24 discrete range bins of size 0.2m and 48 heading bins of size 7.5$\degree$. The RGB panorama frames are encoded with the same ResNet-152 used by the VLN agent. 

A U-Net architecture fuses these RGB features with the obstacle map to predict a radial map of subgoal candidates. This map follows the same range and heading discretization as the input. Gaussian non-maximum suppression filters localized predictions. The $k=5$ candidates with the highest probability mass are converted into $(r, \theta)$ candidates consumed by the VLN agent.

Following Anderson \etal \cite{anderson2020sim}, we train the SGM to minimize Sinkhorn divergence \cite{cuturi2013sinkhorn} to ground-truth subgoal candidates on the VLN nav-graph. Nodes from Val-Unseen are held out for validation. To better match the visual domain of  3D-reconstructed environments, we train the SGM with panoramas rendered using the Habitat Simulator. We provide an ablation analysis motivating our changes to 360$\degree$ laser scanning and reconstructed RGB vision in the supplementary.

%% file: sections/06_results.tex
\csection{Results and Analysis}

We evaluate the transferability of VLN agents to continuous environments using the harness described above. We identify potential sources of performance degradation and evaluate their isolated impacts. Specifically, our analysis starts in VLN and moves toward VLN-CE, covering differences in dataset episodes (\ref{sec:subset}), replacing MP3D panoramas with reconstructed vision (\ref{sec:vision}), using a navigation policy to reach selected subgoals (\ref{sec:nav}), and removing nav-graph subgoals (\ref{sec:graph}). After demonstrating favorable performance over previous agents trained purely in VLN-CE (\ref{sec:leaderboard}), we diagnose a remaining failure mode to be addressed in both future VLN-2-VLNCE transfer and sim-2-real transfer efforts (\ref{sec:elevations}).

\csubsection{How different is the VLN-CE subset of R2R?}
\label{sec:subset}

We find the VLN-CE subset is slightly more difficult but comparable to the full R2R dataset. This suggests that performance in VLN is accurate to, but slightly over-estimates, the upper bound on performance transferred to VLN-CE. We make this comparison in \tabref{tab:subset} by evaluating \vlnbert\ on the topological VLN task for the VLN and VLN-CE subsets of R2R. Differences in these datasets arise from conversion errors which cause the VLN-CE subset of R2R to contain 22\% fewer trajectories than in VLN. \vlnbert\ performs worse on the VLN-CE subset by 3 \texttt{SR} in Val-Seen and 1 \texttt{SR} in Val-Unseen.

\setlength{\tabcolsep}{.275em}
\begin{table}[t]
	\caption{
	Difficulty of Room-to-Room (R2R) episodes in the VLN-CE subset \vs all episodes of R2R. We find performance on the VLN-CE subset is slightly lower than full R2R; Val-Seen success drops by 3 points and Val-Unseen success drops by 1 point.}
	\tightcaption
	\label{tab:subset}
	\begin{center}
	    \resizebox{\textwidth}{!}{
		\begin{tabular}{cl cccsc c cccsc c cccsc}
			\toprule
            \multicolumn{2}{l}{\scriptsize\textbf{Task: VLN}} & \multicolumn{5}{c}{\scriptsize\textbf{\valtrain}} & & \multicolumn{5}{c}{\scriptsize\textbf{Val-Seen}} & & \multicolumn{5}{c}{\scriptsize\textbf{Val-Unseen}} \\
			\cmidrule{3-7} \cmidrule{9-13}
			\cmidrule{15-19}
			\scriptsize \# & \scriptsize Dataset &  \scriptsize\textbf{\texttt{TL}}~$\downarrow$ & \scriptsize\textbf{\texttt{NE}}~$\downarrow$ &
			 \scriptsize\textbf{\texttt{OS}}~$\uparrow$ & \scriptsize\textbf{\texttt{SR}}~$\uparrow$ & \scriptsize\textbf{\texttt{SPL}}~$\uparrow$ & & \scriptsize\textbf{\texttt{TL}}~$\downarrow$ & \scriptsize\textbf{\texttt{NE}}~$\downarrow$ &
			
			\scriptsize\textbf{\texttt{OS}}~$\uparrow$ & \scriptsize\textbf{\texttt{SR}}~$\uparrow$ & \scriptsize\textbf{\texttt{SPL}}~$\uparrow$& & \scriptsize\textbf{\texttt{TL}}~$\downarrow$ & \scriptsize\textbf{\texttt{NE}}~$\downarrow$ &  \scriptsize\textbf{\texttt{OS}}~$\uparrow$ & \scriptsize\textbf{\texttt{SR}}~$\uparrow$ & \scriptsize\textbf{\texttt{SPL}}~$\uparrow$\\
			\midrule
            \scriptsize \texttt{1} & Room-to-Room \cite{anderson2018vision}
                 &  9.98 & 0.93 & 95 & 93 & 90
                && 10.81 & 3.00 & 76 & 72 & 68
                && 11.85 & 4.24 & 67 & 61 & 56 \\
			\scriptsize \texttt{2} & VLN-CE subset \cite{krantz2020beyond}
			     &  9.96 & 1.00 & 94 & 93 & 89
			    && 10.94 & 3.30 & 74 & 69 & 65
			    && 11.79 & 4.43 & 66 & 60 & 54 \\
			\bottomrule
		\end{tabular}}
	\end{center}
\end{table}

\csubsection{What is the visual domain gap from VLN to VLN-CE?}
\label{sec:vision}

\begin{figure}[t]
    \centering
    \includegraphics[width=0.95\textwidth]{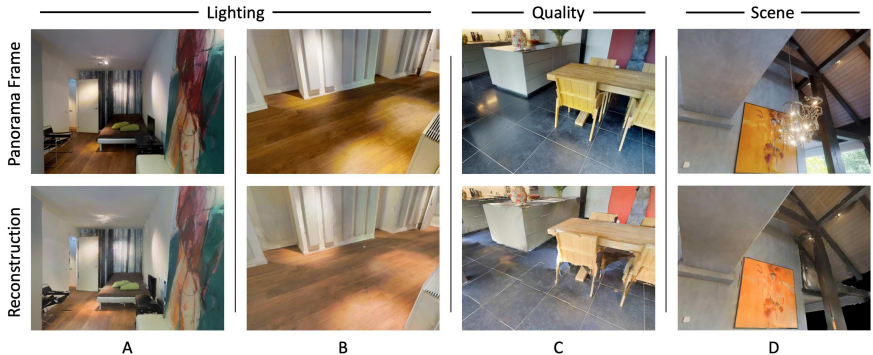}
    \caption{Visual observation differences between VLN and VLN-CE. VLN renders observations from high-quality Matterport3D panoramas (top) and VLN-CE renders observations from 3D scene reconstructions (bottom). Examples of domain differences include lighting (A, B), furniture resolution (C), and object presence in the scene (D).}
    \label{fig:visual}
    \vspace{0pt}
\end{figure}

We find a significant visual domain gap from VLN to VLN-CE in previously-seen environments of 10-13 \texttt{SR} and a moderate gap in novel environments of 3 \texttt{SR} (\tabref{tab:visual}). This suggests that VLN agents may be overfitting to visual representation in training. This visual domain gap exists despite VLN and VLN-CE being derived from the same real-world scenes and sharing paths. We demonstrate characteristic examples in \figref{fig:visual}. VLN observations rendered from Matterport3D panoramas (MP3D panos) are high quality and captured directly from Matterport Pro2 3D cameras \cite{Matterport3D}. On the other hand, VLN-CE observations are rendered from 3D scene reconstructions in Habitat-Sim (Habitat renders) which introduce reconstruction errors and a domain gap in lighting. We demonstrate that training the agent with Habitat renders eliminates this gap almost entirely.

\xhdr{Experiment Setup.}
We evaluate \vlnbert\ in VLN with MP3D panos and Habitat renders. To compute vision features from Habitat, we map each node in VLN to a matching location in VLN-CE using known camera poses. We capture a panorama in the scene reconstruction matching the camera height, angle, and resolution expected in VLN. We then encode the panorama with the same feature encoder used by \vlnbert, a ResNet-152 trained on Places365 \cite{he_cvpr16,zhou2017places}. These features are then used as a drop-in replacement for MP3D pano features.

\xhdr{Zero-Shot Generalization.}
We present zero-shot generalization performance in \tabref{tab:visual}, row 1 \vs 2. Switching to Habitat renders results in success rate drops of 13 points in \valtrain, 10 points in Val-Seen, and 3 points in Val-Unseen. The significant performance degradation in seen environments (\valtrain\ and Val-Seen) demonstrates the impact of the visual domain gap between VLN and VLN-CE. The relatively smaller drop in Val-Unseen suggests that this gap is not as catastrophic when generalizing to novel environments and that VLN agents may be overfitting to visual representation in training environments.

\xhdr{Domain Transfer.}
We show in row 4 that re-training \vlnbert\ directly on reconstructed vision can entirely recover task performance in seen environments and recover all by 1 point in success in Val-Unseen (row 4 \vs 1). This reconstruction-trained model achieves a success rate of 60 in Val-Unseen regardless of using MP3D panos or Habitat renders (rows 3 and 4). This suggests that training VLN agents on Habitat renders may result in less visual overfitting.

\setlength{\tabcolsep}{.275em}
\begin{table}[t]
	\caption{Effect of the visual domain gap between Matterport3D panoramas (MP3D) and reconstruction renders (Recon.) on VLN performance. Reconstructed vision decreases success rate in Val-Unseen by 3 points (rows 1 \vs 2). This drop is mitigated by training VLN agents directly on reconstructed images (row 4).}
	\tightcaption
	\label{tab:visual}
	\begin{center}
	    \resizebox{\textwidth}{!}{
		\begin{tabular}{cll c cccsc c cccsc c cccsc}
			\toprule
            \multicolumn{1}{l}{\scriptsize\textbf{Task: VLN}} &
            \multicolumn{2}{c}{\scriptsize{Camera}} &
            \multicolumn{5}{c}{\scriptsize\textbf{\valtrain}} & & \multicolumn{5}{c}{\scriptsize\textbf{Val-Seen}} & & \multicolumn{5}{c}{\scriptsize\textbf{Val-Unseen}} \\
            
			\cmidrule{2-3}
			\cmidrule{5-9} \cmidrule{11-15} \cmidrule{17-21}
			\scriptsize \# & \multicolumn{1}{c}{\scriptsize Train} & \multicolumn{1}{c}{\scriptsize Eval} && \scriptsize\textbf{\texttt{TL}}~$\downarrow$ & \scriptsize\textbf{\texttt{NE}}~$\downarrow$ &
			 \scriptsize\textbf{\texttt{OS}}~$\uparrow$ & \scriptsize\textbf{\texttt{SR}}~$\uparrow$ & \scriptsize\textbf{\texttt{SPL}}~$\uparrow$ & & \scriptsize\textbf{\texttt{TL}}~$\downarrow$ & \scriptsize\textbf{\texttt{NE}}~$\downarrow$ &
			
			\scriptsize\textbf{\texttt{OS}}~$\uparrow$ & \scriptsize\textbf{\texttt{SR}}~$\uparrow$ & \scriptsize\textbf{\texttt{SPL}}~$\uparrow$& & \scriptsize\textbf{\texttt{TL}}~$\downarrow$ & \scriptsize\textbf{\texttt{NE}}~$\downarrow$ &  \scriptsize\textbf{\texttt{OS}}~$\uparrow$ & \scriptsize\textbf{\texttt{SR}}~$\uparrow$ & \scriptsize\textbf{\texttt{SPL}}~$\uparrow$\\
			\midrule
            \scriptsize \texttt{1} & \multirow{2}{*}{\shortstack[l]{MP3D}} & MP3D &
                 &  9.98 & 0.93 & 95 & 93 & 90
                && 10.81 & 3.00 & 76 & 72 & 68
                && 11.85 & 4.24 & 67 & 61 & 56 \\
			\scriptsize \texttt{2} & & Recon. &
			     & 11.17 & 2.20 & 84 & 80 & 76  
			    && 11.66 & 4.02 & 68 & 62 & 58
			    && 11.69 & 4.56 & 64 & 58 & 53 \\
		    \cmidrule{1-21}
            \scriptsize \texttt{3} & \multirow{2}{*}{\shortstack[l]{Recon.}} & MP3D &
                 & 10.63 & 1.18 & 92 & 90 & 86
                && 11.35 & 3.41 & 72 & 66 & 61
                && 11.72 & 4.12 & 66 & 60 & 54 \\
			\scriptsize \texttt{4} & & Recon. &
			     &  9.76 & 0.48 & 98 & 98 & 96
			    && 10.56 & 2.88 & 77 & 72 & 69
			    && 11.16 & 4.14 & 66 & 60 & 56 \\
			\bottomrule
		\end{tabular}}
	\end{center}
\end{table}

\csubsection{What is the navigation gap between VLN and VLN-CE?}
\label{sec:nav}

Conveying the VLN agent to selected subgoals using a realistic navigator causes a decrease in performance across all splits, highlighted by a 5 point drop in Val-Unseen success. The VLN task, which effectively teleports between subgoals, makes navigation assumptions of perfect obstacle avoidance and stopping precision. These challenges must be contended with in VLN-CE. In the worst case, navigation failure causes episode failure regardless of the decisions made by the VLN agent. In the best case, an unexpected pose and observation are produced at the next time step, a situation VLN agents do not encounter in training.

\xhdr{Experiment Setup.}
We evaluate our reconstruction-trained \vlnbert\ in VLN-CE with each navigation policy from \secref{sec:nav_agents} and provide known subgoal candidates from the VLN nav-graph. We note that navigation errors introduce ambiguity in graph localization which must be resolved to provide the next subgoal candidates. In our experiments, we keep the agent at its navigation-terminated location but provide subgoal candidates by assuming it is located at the nearest node on the nav-graph by geodesic distance. This reflects the consequences of poor navigation where a large navigation error may cause the VLN agent to observe candidates from a different graph node than intended. We first evaluate the impact of using the VLN-CE action space instead of teleportation. We then evaluate the effect of a realistic navigation policy. Results of both comparisons are in \tabref{tab:navigator}.

\xhdr{Teleportation \vs VLN-CE Actions.} We use the Oracle Policy to represent ideal actions within the VLN-CE action space. We find that performance under the Oracle Policy matches performance under Teleportation in Val-Unseen (row 2 \vs 1). This suggests that the navigation gap between VLN and VLN-CE can be minimized with performant navigation policies. This is an expected result due to the navigation verification that originally culled episodes for the VLN-CE dataset (see \cite{krantz2020beyond}). We additionally record the navigation error of the Oracle Policy for each short-range navigation performed, finding an average navigation error of 0.11m in Val-Seen and 0.08m in Val-Unseen. This demonstrates that the VLN agent is robust to position jitter about official graph nodes.

\xhdr{Optimal Navigation \vs Realistic Navigation.}
In row 3, we present performance under our Local Policy. Compared to the Oracle Policy, we find a drop in success in both Val-Seen (3 points) and Val-Unseen (5 points). The SPL drop is identical, suggesting that path efficiency remains high in successful episodes. The average short-range navigation error is 0.26m in Val-Seen and 0.40m in Val-Unseen, which are higher than that of the Oracle Policy. We observe that most calls to the Local Policy perform similarly to the Oracle Policy. However, the Local Policy has a longer tail of significant navigation failures (error $>$0.5m). Specifically, the Local Policy has a $>$0.5m failure rate that is 15x higher than the Oracle Policy. We expand on this error comparison in the supplementary.

\setlength{\tabcolsep}{.275em}
\begin{table}[t]
	\caption{Impact of navigation policies on VLN agents operating in VLN-CE. We assume a known nav-graph but require lower-level actions between nodes. The Oracle Policy results in better performance than the Local Policy, with a 5 \texttt{SR} Val-Unseen gap.}
	\tightcaption
	\label{tab:navigator}
	\begin{center}
	    \resizebox{\textwidth}{!}{
		\begin{tabular}{cl cccsc c cccsc c cccsc}
			\toprule
			\multicolumn{2}{l}{\scriptsize\textbf{Task: VLN-CE}} & \multicolumn{5}{c}{\scriptsize\textbf{\valtrain}} & & \multicolumn{5}{c}{\scriptsize\textbf{Val-Seen}} & & \multicolumn{5}{c}{\scriptsize\textbf{Val-Unseen}} \\
			\cmidrule{3-7} \cmidrule{9-13}
			\cmidrule{15-19}
			\scriptsize \# & \scriptsize Navigator &  \scriptsize\textbf{\texttt{TL}}~$\downarrow$ & \scriptsize\textbf{\texttt{NE}}~$\downarrow$ &
			 \scriptsize\textbf{\texttt{OS}}~$\uparrow$ & \scriptsize\textbf{\texttt{SR}}~$\uparrow$ & \scriptsize\textbf{\texttt{SPL}}~$\uparrow$ & & \scriptsize\textbf{\texttt{TL}}~$\downarrow$ & \scriptsize\textbf{\texttt{NE}}~$\downarrow$ &
			
			\scriptsize\textbf{\texttt{OS}}~$\uparrow$ & \scriptsize\textbf{\texttt{SR}}~$\uparrow$ & \scriptsize\textbf{\texttt{SPL}}~$\uparrow$& & \scriptsize\textbf{\texttt{TL}}~$\downarrow$ & \scriptsize\textbf{\texttt{NE}}~$\downarrow$ &  \scriptsize\textbf{\texttt{OS}}~$\uparrow$ & \scriptsize\textbf{\texttt{SR}}~$\uparrow$ & \scriptsize\textbf{\texttt{SPL}}~$\uparrow$\\
			\midrule
            \scriptsize \texttt{1} & Teleportation
                 &  10.04 & 0.58 & 97 & 97 & 88
                && 11.28 & 3.24 & 75 & 70 & 63
                && 11.98 & 4.06 & 66 & 60 & 52 \\
			\scriptsize \texttt{2} & Oracle Policy
			     &  10.09 & 0.72 & 97 & 96 & 87
			    && 11.19 & 3.10 & 75 & 69 & 61
			    && 12.04 & 4.07 & 68 & 60 & 52 \\
			\scriptsize \texttt{3} & Local Policy
			     &  10.58 & 1.57 & 90 & 88 & 78
			    && 11.37 & 3.49 & 72 & 66 & 58
			    && 12.28 & 4.51 & 63 & 55 & 47 \\
			\bottomrule
		\end{tabular}}
	\end{center}
\end{table}

\setlength{\tabcolsep}{.275em}
\begin{table}[t]
	\caption{Comparing VLN-CE performance when subgoal candidates are provided by the nav-graph (rows 1,2) \vs predicted by a subgoal generation module (SGM)(rows 3,4). Rows 3 and 4 demonstrate a complete harness for a VLN agent admissible in VLN-CE, consisting of both subgoal candidate generation and lower-level navigation.
	}
	\tightcaption
	\label{tab:graph}
	\begin{center}
	    \resizebox{\textwidth}{!}{
		\begin{tabular}{clc c cccsc c cccsc c cccsc}
			\toprule
            \multicolumn{2}{l}{\scriptsize\textbf{Task: VLN-CE}} & & & \multicolumn{5}{c}{\scriptsize\textbf{\valtrain}} & & \multicolumn{5}{c}{\scriptsize\textbf{Val-Seen}} & & \multicolumn{5}{c}{\scriptsize\textbf{Val-Unseen}} \\
            
			\cmidrule{5-9} \cmidrule{11-15} \cmidrule{17-21}
			\scriptsize \# & \multicolumn{1}{c}{\scriptsize Subgoal Candidates} & \multicolumn{1}{c}{\scriptsize F-tune} && \scriptsize\textbf{\texttt{TL}}~$\downarrow$ & \scriptsize\textbf{\texttt{NE}}~$\downarrow$ &
			 \scriptsize\textbf{\texttt{OS}}~$\uparrow$ & \scriptsize\textbf{\texttt{SR}}~$\uparrow$ & \scriptsize\textbf{\texttt{SPL}}~$\uparrow$ & & \scriptsize\textbf{\texttt{TL}}~$\downarrow$ & \scriptsize\textbf{\texttt{NE}}~$\downarrow$ &
			
			\scriptsize\textbf{\texttt{OS}}~$\uparrow$ & \scriptsize\textbf{\texttt{SR}}~$\uparrow$ & \scriptsize\textbf{\texttt{SPL}}~$\uparrow$& & \scriptsize\textbf{\texttt{TL}}~$\downarrow$ & \scriptsize\textbf{\texttt{NE}}~$\downarrow$ &  \scriptsize\textbf{\texttt{OS}}~$\uparrow$ & \scriptsize\textbf{\texttt{SR}}~$\uparrow$ & \scriptsize\textbf{\texttt{SPL}}~$\uparrow$\\
			\midrule
            \scriptsize \texttt{1} & Nav-Graph & - &
			     & 10.58 & 1.57 & 90 & 88 & 78
			    && 11.37 & 3.49 & 72 & 66 & 58
			    && 12.28 & 4.51 & 63 & 55 & 47 \\
			\scriptsize \texttt{2} & Optimal SGM & - &
			     & 11.40 & 2.07 & 86 & 81 & 72  
			    && 12.69 & 3.78 & 71 & 63 & 55
			    && 14.56 & 4.96 & 61 & 49 & 41 \\
		    \cmidrule{1-21}
            \scriptsize \texttt{3} & \multirow{2}{*}{SGM} & - &
                 & 13.12 & 3.54 & 71 & 65 & 55
                && 12.69 & 4.51 & 60 & 51 & 44
                && 13.74 & 5.83 & 51 & 41 & 35 \\
			\scriptsize \texttt{4} & & \checkmark &
			     &  11.15 & 3.77 & 73 & 66 & 56
			    && 11.18 & 4.67 & 61 & 52 & 44
			    && 10.69 & 6.07 & 52 & 43 & 36 \\
			\bottomrule
		\end{tabular}}
	\end{center}
\end{table}

\csubsection{What is the subgoal candidate gap?}
\label{sec:graph}

In the previous experiments, VLN agents were selecting from subgoal candidates provided by the nav-graph, but these are not available in VLN-CE. Here, we predict subgoal candidates online (at each time step) and observe significant performance degradation (14-23 \texttt{SR}). Fine-tuning the VLN agent on the online distribution of subgoal candidates slightly mitigates this drop, recovering 1-2 \texttt{SR}.

\xhdr{Experiment Setup.} We replace the nav-graph with our subgoal generation module (SGM: \secref{sec:sgm}) as the source of subgoal candidates. We join our SGM with reconstruction-trained \vlnbert\ and our Local Policy as navigator. In \tabref{tab:graph}, we evaluate the impact of the prediction space and the SGM predictions.

\xhdr{Optimal SGM Predictions.}
We first consider the case where SGM predictions optimally match the nav-graph (\tabref{tab:graph}, row 2). We convert the nav-graph subgoals into a radial prediction map that matches the SGM output space. This discretizes the subgoal candidate locations into polar bins with a 0.20m range and 7.5$\degree$ resolution. This prediction space accounts for a non-trivial drop in performance, resulting in a 3 point drop in Val-Seen success rate and a 6 point drop in Val-Unseen success rate (row 2 \vs 1).

\xhdr{Performance with SGM.}
Row 3 demonstrates subgoal predictions from the SGM with no nav-graph reliance. This experiment is compliant with the VLN-CE task definition; subgoal candidates are predicted from egocentric observations, a VLN agent selects a candidate, and a local policy conveys the agent using VLN-CE actions. We find a decrease of 12 \texttt{SR} in Val-Seen and 8 \texttt{SR} in Val-Unseen. These drops indicate a large domain gap between the nav-graph and the SGM. This result parallels findings in sim-2-real transfer by Anderson \etal \cite{anderson2020sim} that point to an alignment problem between the VLN nav-graph and the SGM.

\xhdr{Fine-Tuning \vlnbert\ with SGM.}
We seek to determine if fine-tuning a VLN agent on the SGM candidates can reclaim nav-graph performance. To motivate, consider how learned priors in VLN could result in detrimental behavior under a different distribution of subgoals. For example, a slight change in the distribution of subgoal distances could break a learned time horizon prior -- the VLN agent may choose to stop either too early or too late. Such a prior can be very strong in VLN since all paths require 5-7 actions. We fine-tune our reconstruction-trained \vlnbert\ via imitation learning in VLN-CE. Specifically, we train with teacher forcing to maximize the probability of predicting the subgoal candidate that greedily minimizes the geodesic distance to the goal. We set a batch size of 12, a learning rate of 1e-7, and train with cross-entropy loss and early stopping. In \tabref{tab:graph} row 4, we present the result of fine-tuning on SGM predictions. Performance increases slightly across all validation splits, highlighted by a 2-point increase in Val-Unseen success rate. This still leaves a 6 point gap in success rate to predicting optimal subgoal candidates and a 12 point gap in success rate to candidates directly from the nav-graph. Fine-tuning experiments such as this one cannot be performed on the rigid topology of VLN. Neither can they be performed practically in reality, making VLN-CE a promising setting for generalizing VLN agents for sim-2-sim or sim-2-real transfer.

\csubsection{Comparison With Previous VLN-CE Models}
\label{sec:leaderboard}

We compare our best agent against previously published methods on the VLN-CE Challenge Leaderboard\footnote{\href{https://eval.ai/web/challenges/challenge-page/719/overview}{eval.ai/web/challenges/challenge-page/719}}. This agent includes subgoal candidates generated by the SGM, our reconstruction-trained \vlnbert\ fine-tuned on SGM candidates, and Local Policy navigation. We label our submission \vlncebert. As shown in \tabref{tab:existing}, \vlncebert\ surpasses the success rate of all previous models on all splits including Test. Notably, all previous methods are trained exclusively in VLN-CE, with the best model amongst them requiring extensive compute (7000+ GPU-hours) to achieve a 32 \texttt{SR}. With just 12 GPU-hours of fine-tuning, our model surpasses this performance level by 12 points \texttt{SR} to achieve a 44 \texttt{SR} in Test (a relative improvement of 38\%). This result shows that VLN to VLN-CE transfer is a highly promising avenue for making progress on the VLN-CE task and closing the gap to VLN.

\setlength{\tabcolsep}{.275em}
\begin{table}[t]
	\caption{Results on the VLN-CE Challenge Leaderboard. Our submission outperforms previously-published results, demonstrating a 12 point improvement over the next best model in success rate on Test (a 38\% relative improvement).}
	\tightcaption
	\label{tab:existing}
	\begin{center}
	    \resizebox{\textwidth}{!}{
		\begin{tabular}{cl cccsc c cccsc c cccsc}
			\toprule
            \multicolumn{2}{l}{\scriptsize\textbf{Task: VLN-CE}} & \multicolumn{5}{c}{\scriptsize\textbf{Val-Seen}} & & \multicolumn{5}{c}{\scriptsize\textbf{Val-Unseen}} & & \multicolumn{5}{c}{\scriptsize\textbf{Test}} \\
			\cmidrule{3-7} \cmidrule{9-13} \cmidrule{15-19}
			\scriptsize \# & \scriptsize Model &  \scriptsize\textbf{\texttt{TL}}~$\downarrow$ & \scriptsize\textbf{\texttt{NE}}~$\downarrow$ &
			 \scriptsize\textbf{\texttt{OS}}~$\uparrow$ & \scriptsize\textbf{\texttt{SR}}~$\uparrow$ & \scriptsize\textbf{\texttt{SPL}}~$\uparrow$ & & \scriptsize\textbf{\texttt{TL}}~$\downarrow$ & \scriptsize\textbf{\texttt{NE}}~$\downarrow$ &
			
			\scriptsize\textbf{\texttt{OS}}~$\uparrow$ & \scriptsize\textbf{\texttt{SR}}~$\uparrow$ & \scriptsize\textbf{\texttt{SPL}}~$\uparrow$& & \scriptsize\textbf{\texttt{TL}}~$\downarrow$ & \scriptsize\textbf{\texttt{NE}}~$\downarrow$ &  \scriptsize\textbf{\texttt{OS}}~$\uparrow$ & \scriptsize\textbf{\texttt{SR}}~$\uparrow$ & \scriptsize\textbf{\texttt{SPL}}~$\uparrow$\\
			\midrule
            \scriptsize \texttt{1} & \textbf{\vlncebert} (ours)
                 & 11.18 & \textbf{4.67} & \textbf{61} & \textbf{52} & \textbf{44}
			    && 10.69 & \textbf{6.07} & \textbf{52} & \textbf{43} & \textbf{36}
                && 11.43 & \textbf{6.17} & \textbf{52} & \textbf{44} & \textbf{37} \\
			\scriptsize \texttt{2} & HPN+DN \cite{krantz2021waypoint}
			     & 8.54 & 5.48 & 53 & 46 & 43
			    && 7.62 & 6.31 & 40 & 36 & 34 
			    && 8.02 & 6.65 & 37 & 32 & 30 \\
			\scriptsize \texttt{3} & WPN+DN \cite{krantz2021waypoint}
			     & 9.52 & 6.23 & 45 & 37 & 33 
			    && 9.86 & 6.93 & 40 & 33 & 29
		        && 9.68 & 7.49 & 36 & 29 & 25 \\
			\scriptsize \texttt{4} & LAW \cite{raychaudhuri2021language}
			     & 9.34 & 6.35 & 49 & 40 & 37
		        && 8.89 & 6.83 & 44 & 35 & 31
		        && 9.67 & 7.69 & 38 & 28 & 25 \\
			\scriptsize \texttt{5} & CMA+PM+DA+Aug \cite{krantz2020beyond}
			     & 9.06 & 7.21 & 44 & 34 & 32
		        && 8.27 & 7.60 & 36 & 29 & 27
		        && 8.85 & 7.91 & 36 & 28 & 25 \\
			\bottomrule
		\end{tabular}}
	\end{center}
\end{table}

\begin{figure}[t]
    \centering
    \begin{minipage}{0.79\textwidth}
    \subfigure[\scriptsize Greedy Oracle  \label{fig:elev_greedy}]{
        \includegraphics[width=0.32\textwidth]{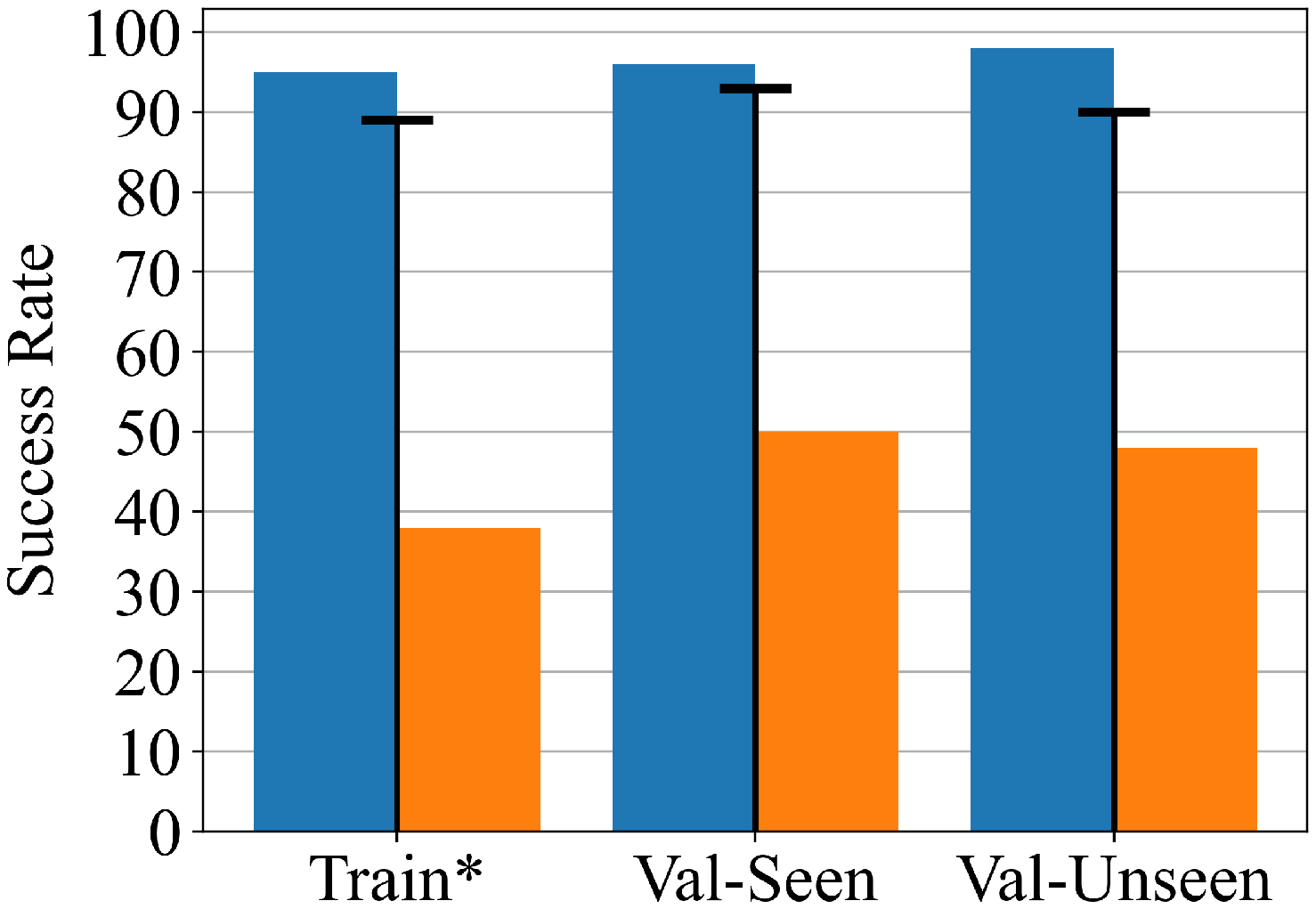}}\hfill
    \subfigure[\scriptsize \vlncebert \label{fig:elev_rt}]{
        \includegraphics[width=0.32\textwidth]{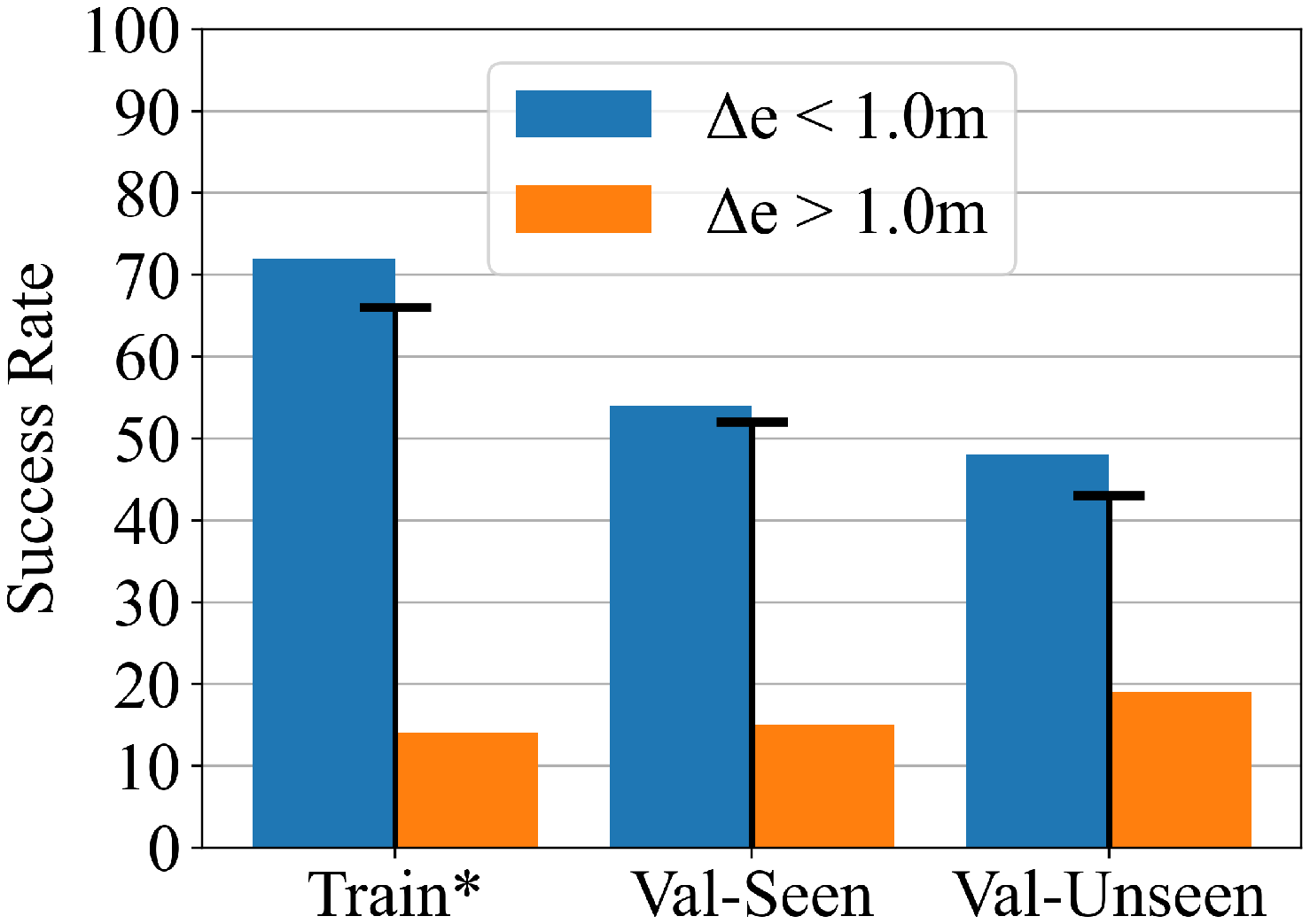}}\hfill
    \subfigure[\scriptsize VLN: rt-\vlnbert \label{fig:elev_rt_vln}]{
        \includegraphics[width=0.32\textwidth]{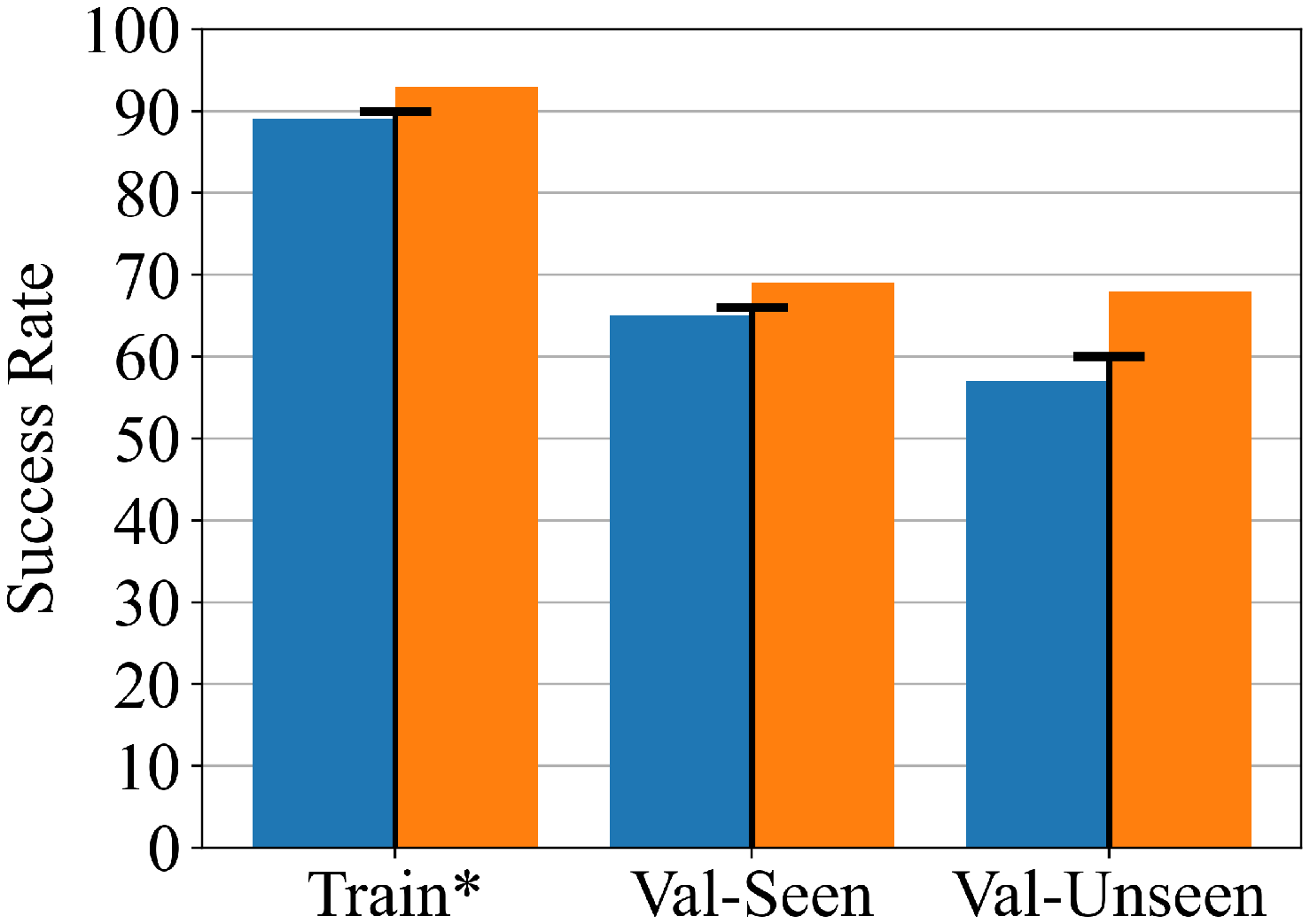}}\hfill
    \end{minipage}
    \hfill
	\begin{minipage}{0.2\textwidth}
	    \subfigure[\scriptsize Episodes  \label{fig:elev_table}]{
    	   \rule[12pt]{0pt}{25pt}
    	    \resizebox{0.99\textwidth}{!}{
    	        \renewcommand{\arraystretch}{1.4}
                \begin{tabular}{l cc}
                    \toprule
                    & \multicolumn{2}{c}{\scriptsize\textbf{$\Delta$e $>$1.0m (\%)}} \\
                    Split & VLN-CE & VLN \\
                    \midrule
                    Train      &  8.8 & 19.6 \\
                    \valtrain  & 11.0 & 23.2 \\
                    Val-Seen   &  6.2 & 15.9 \\
                    Val-Unseen & 16.0 & 21.5 \\
                    \bottomrule
                \end{tabular}
    	    }
    	    \rule[-35pt]{0pt}{35pt}
        }
    \end{minipage}
    \caption{Success rates in episodes with high (>1.0m) \vs low (<1.0m) elevation delta. Black bars indicate success rate irrespective of elevation. (a,b) use SGM subgoal candidates in VLN-CE and perform worse with elevation. However in VLN (c), our reconstruction-trained agent performs better with elevation. (d) shows the percentage of episodes that have a $>$1.0m delta.
    }
    \label{fig:elev}
\end{figure}

\csubsection{Failures of Subgoal Candidate Generation}
\label{sec:elevations}

Starting from \vlnbert\ operating in VLN, we diagnosed a 1 \texttt{SR} drop from dataset differences, a 1 \texttt{SR} drop from the visual domain gap, a 5 \texttt{SR} drop from navigation policies, and a 12 \texttt{SR} drop from subgoal candidate generation. We now look into what failure modes constitute the subgoal candidate drop. We find that half of these failures (6/12 \texttt{SR}) are caused by episodes that require the agent to gain or lose at least 1.0m of elevation (\ie traversing stairs). In such episodes, an oracle VLN agent (a greedy-optimal subgoal selector) paired with the Oracle Policy for navigation is successful less than 50\% of the time. This indicates that the SGM has a recall problem in generating elevation-changing subgoals necessary for success. Details and experimental support are below.

\xhdr{Oracle Selection and Navigation.}
We first consider whether candidates produced by the SGM can lead to episode success independent of failures stemming from selection or navigation issues. We define a greedy oracle VLN agent that selects the subgoal candidate with minimal geodesic distance to the goal. \texttt{STOP} is called when no candidate is closer to the goal than the current position. We run this oracle over SGM candidates and navigate via the Oracle Policy. We find success rates of 89\% in \valtrain, 93\% in Val-Seen, and 90\% in Val-Unseen (\figref{fig:elev_greedy}). However, failures disproportionately stem from episodes requiring elevation change. This is exemplified in \valtrain\ where episodes requiring at least a 1.0m elevation change result in 38\% success -- a highly restrictive upper bound.

\xhdr{Elevation-Based Performance of \vlncebert.}
We compare success rates for \vlncebert\ in episodes requiring a high (>1.0m) \vs low (<1.0m) elevation delta (\figref{fig:elev_rt}). In Val-Unseen, there is a 29 \texttt{SR} gap between such episodes (19 \texttt{SR} \vs 48 \texttt{SR}). This is unsurprising given the poor performance of the oracle VLN agent with high elevation change. The failure of the SGM to predict elevation subgoals (a recall problem) may be related to how it is trained; subgoal candidate locations are highly correlated with free-space in the 2D laser scan. However 2D free-space is a poor predictor of subgoals with elevation change, which are less represented in the training data (\figref{fig:elev_table}).

\xhdr{Elevation-Based Performance in VLN.}
Elevation-based VLN evaluation provides an upper bound on VLN-CE performance in elevation episodes. Surprisingly, we find in \figref{fig:elev_rt_vln} that our reconstruction-trained \vlnbert\ has a higher success rate in episodes requiring a high elevation delta (68 \texttt{SR} \vs 57 \texttt{SR} in Val-Unseen). We suspect this result stems from episodes that feature stairs have a smaller search space; nav-graph nodes on stairs have a smaller degree. From this result, we estimate that solving the elevation problem in VLN-CE can mitigate 6 points of the 12 \texttt{SR} drop attributable to subgoal candidate generation, leaving a cumulative 12 \texttt{SR} gap between VLN and VLN-CE.

\csubsection{Qualitative Example.}
\label{sec:qualitative}

\begin{figure}[t]
    \centering
    \includegraphics[width=\textwidth]{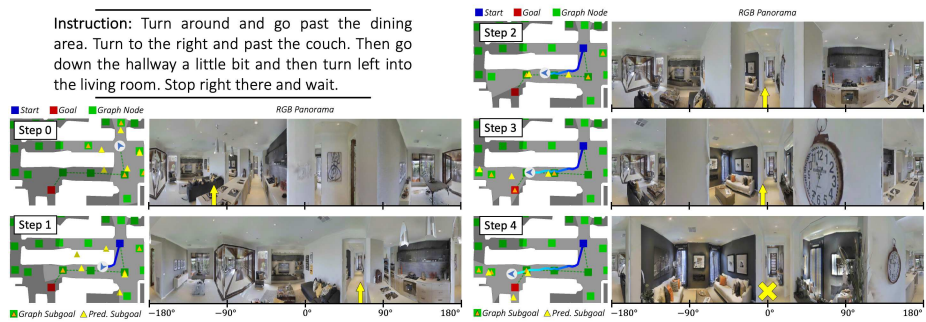}
    \caption{Qualitative example of our \vlncebert\ agent operating in VLN-CE.}
    \label{fig:qual}
\end{figure}

We present a qualitative example of \vlncebert\ successfully navigating in a novel environment within VLN-CE (\figref{fig:qual}). This example demonstrates all components of our agent: subgoal candidate generation, subgoal selection, and navigation. In each step, the subgoal candidates (large, yellow triangles) enable navigation in primary directions. However, not all candidates can be reached, like in Step 1 where a candidate is centered on the kitchen countertop. The SGM candidates diverge from nav-graph node locations. This leads the VLN agent to observe the environment from poses off the nav-graph, yet proper candidate selections are still made over the course of 5 actions. Notice that between steps 0 and 1, the selected subgoal cannot be navigated to in a straight line. Our Local Policy demonstrates obstacle avoidance while planning and executing a path that traverses around the countertop. This example demonstrates that through modular construction, VLN agents can operate successfully and efficiently in continuous environments. We provide additional examples in the supplementary.

%% file: sections/07_conclusion.tex
\csection{Conclusion}

In summary, we explore the sim-to-sim transfer of instruction-following agents from topological VLN to unconstrained VLN-CE. Our transfer results in an absolute improvement of 12\% over the prior state-of-the-art in VLN-CE. We diagnose the remaining VLN-to-VLNCE gap, identifying subgoal candidate generation as a primary hindrance to transfer. We outline the problem of generating candidates in multi-floor environments to guide future work. Operating VLN agents in continuous environments enables a new interplay between language and navigation topologies. This can lead not only to higher performance in realistic environments, but also to the development of more robust topological navigators.

%% file: sections/08_acknowledgement.tex
\vspace{0.05in}
{
\footnotesize
\xhdr{Acknowledgements.}
This work was supported in part by the DARPA Machine Common Sense program. The views and conclusions contained herein are those of the authors and should not be interpreted as necessarily representing the official policies or endorsements, either expressed or implied, of the U.S. Government, or any sponsor.
}

%% file: sections/09_supp.tex
\csection{Supplementary}

\csubsection{\vlnbert: Validation Performance}

We use the \vlnbert\ model in VLN experiments to evaluate the impact of dataset differences between VLN and VLN-CE (\secref{sec:subset}) and MP3D \vs reconstructed vision (\secref{sec:vision}). For consistency across our experiments, we retrain \vlnbert\ using the official codebase\footnote{\href{https://github.com/YicongHong/Recurrent-VLN-BERT}{github.com/YicongHong/Recurrent-VLN-BERT}}. We train with the (init. PREVALENT) backbone. Our re-trained version performs at 2 \texttt{SR} and 1 \texttt{SPL} lower in Val-Unseen than the published result in \cite{hong2021vln} but matches performance in Val-Seen \texttt{SR} and \texttt{SPL}. We repeat training and evaluation with 3 different random seeds and find performance consistent with what we present in \tabref{tab:subset} and \tabref{tab:visual}.

\csubsection{Oracle Policy Detail}

We use the Oracle Policy with known subgoal candidates that are specified in 3D coordinates. We note that using this oracle with 2D subgoal predictions requires projecting the target location from 2D to 3D. We define a projection procedure $P: (r, \theta) \rightarrow (x,y,z)$ that maps distance and relative heading to global 3D coordinates. In this procedure, the agent's current pose is used to project $(r, \theta)$ to global 2D coordinates $(\hat{x},\hat{z})$. We assume the target exists at the elevation of the agent's pose, $\hat{y}$. Finally, we snap the resulting 3D coordinates $(\hat{x},\hat{y},\hat{z})$ to the nearest position $(x,y,z)$ that exists on the navigation mesh and that has a finite geodesic distance from the agent's current position.

\csubsection{Subgoal Module Ablations}

In \tabref{tab:sgm_ablation} we evaluate ablations of the subgoal generation module (SGM) in VLN-CE. In row 1 \vs 2, we find that training the SGM with Habitat-rendered vision (Recon.) leads to better downstream performance in VLN-CE than MP3D panoramas across all splits (3 \texttt{SR} Val-Unseen, 5 \texttt{SR} Val-Seen, 6 \texttt{SR} \valtrain). We further ablate the 360$\degree$ laser scan to 270$\degree$  and observe an additional performance drop of 5 \texttt{SR} in Val-Unseen, 4 \texttt{SR} in Val-Seen, and 8 \texttt{SR} in \valtrain. Altogether, our modifications of reconstructed vision and 360$\degree$ scanning improve performance under the SGM by 8 \texttt{SR} in Val-Unseen over the SGM proposed in \cite{anderson2020sim}.

\setlength{\tabcolsep}{.275em}
\begin{table}[t]
	\caption{Ablations against the subgoal generation module (SGM). Results are in VLN-CE with reconstruction-trained \vlnbert\ and Local Policy navigator. Row 1 matches \tabref{tab:graph} row 3 in the main paper. Row 2 ablates training with reconstructed vision (Recon.) and row 3 ablates both Recon. and 360$\degree$ laser scan to match the SGM used by Anderson \etal \cite{anderson2020sim}.}
	\tightcaption
	\label{tab:sgm_ablation}
	\begin{center}
	    \resizebox{\textwidth}{!}{
		\begin{tabular}{clcc c cccsc c cccsc c cccsc}
			\toprule
            \multicolumn{2}{l}{\scriptsize\textbf{Task: VLN-CE}} & & & & \multicolumn{5}{c}{\scriptsize\textbf{\valtrain}} & & \multicolumn{5}{c}{\scriptsize\textbf{Val-Seen}} & & \multicolumn{5}{c}{\scriptsize\textbf{Val-Unseen}} \\
            
			\cmidrule{6-10} \cmidrule{12-16} \cmidrule{18-22}
			\scriptsize \# & \multicolumn{1}{c}{\scriptsize Subgoal Candidates} & \multicolumn{1}{c}{\scriptsize Vision} & HFOV && \scriptsize\textbf{\texttt{TL}}~$\downarrow$ & \scriptsize\textbf{\texttt{NE}}~$\downarrow$ &
			 \scriptsize\textbf{\texttt{OS}}~$\uparrow$ & \scriptsize\textbf{\texttt{SR}}~$\uparrow$ & \scriptsize\textbf{\texttt{SPL}}~$\uparrow$ & & \scriptsize\textbf{\texttt{TL}}~$\downarrow$ & \scriptsize\textbf{\texttt{NE}}~$\downarrow$ &
			
			\scriptsize\textbf{\texttt{OS}}~$\uparrow$ & \scriptsize\textbf{\texttt{SR}}~$\uparrow$ & \scriptsize\textbf{\texttt{SPL}}~$\uparrow$& & \scriptsize\textbf{\texttt{TL}}~$\downarrow$ & \scriptsize\textbf{\texttt{NE}}~$\downarrow$ &  \scriptsize\textbf{\texttt{OS}}~$\uparrow$ & \scriptsize\textbf{\texttt{SR}}~$\uparrow$ & \scriptsize\textbf{\texttt{SPL}}~$\uparrow$\\
			\midrule
            \scriptsize \texttt{1} & \multirow{3}{*}{SGM} & Recon. & 360 &
                 & 13.12 & 3.54 & 71 & 65 & 55
                && 12.69 & 4.51 & 60 & 51 & 44
                && 13.74 & 5.83 & 51 & 41 & 35 \\
			\scriptsize \texttt{2} & & MP3D & 360 &
			     & 10.08 & 3.52 & 64 & 59 & 54
			    && 10.41 & 4.80 & 52 & 46 & 41
			    && 10.07 & 5.61	& 45 & 38 & 34 \\
			\scriptsize \texttt{3} & & MP3D & 270 &
			     & 11.24 & 4.14 & 59 & 51 & 46
		        && 10.73 & 5.02 & 47 & 42 & 37
		        && 10.62 & 5.97 & 41 & 33 & 29 \\
			\bottomrule
		\end{tabular}}
	\end{center}
\end{table}

\setlength{\tabcolsep}{.275em}
\begin{table}[t]
	\caption{Short-range navigation errors (subgoal-to-subgoal) of navigation policies. Evaluated while \vlnbert\ is performing the VLN-CE task with nav-graph subgoals (\tabref{tab:navigator}). Navigation errors are in meters and thresholded error values (>Xm) are reported as a percent of all short-range navigations.}
	\tightcaption
	\label{tab:nav_errors}
	\begin{center}
	    \resizebox{\textwidth}{!}{
		\begin{tabular}{cl sccc c sccc c sccc}
			\toprule
			\multicolumn{2}{l}{\scriptsize\textbf{Task: VLN-CE}} & \multicolumn{4}{c}{\scriptsize\textbf{\valtrain}} & & \multicolumn{4}{c}{\scriptsize\textbf{Val-Seen}} & & \multicolumn{4}{c}{\scriptsize\textbf{Val-Unseen}} \\
			\cmidrule{3-6}
			\cmidrule{8-11}
			\cmidrule{13-16}
			\scriptsize \# & \scriptsize Navigator
		    & \scriptsize\textbf{\texttt{NE}}~$\downarrow$ & \scriptsize\textbf{>0.2m} & \scriptsize\textbf{>0.5m} & \scriptsize\textbf{>1.0m}
		    
		    & & \scriptsize\textbf{\texttt{NE}}~$\downarrow$ & \scriptsize\textbf{>0.2m} & \scriptsize\textbf{>0.5m} & \scriptsize\textbf{>1.0m}
		    
		    & & \scriptsize\textbf{\texttt{NE}}~$\downarrow$ & \scriptsize\textbf{>0.2m} & \scriptsize\textbf{>0.5m} & \scriptsize\textbf{>1.0m}\\
			\midrule
			\scriptsize \texttt{1} & Oracle Policy
    		      & 0.09 & 0.24 & 0.24 & 0.24
    		     && 0.11 & 0.74 & 0.69 & 0.69
    		     && 0.08 & 0.53 & 0.42 & 0.29\\
			\scriptsize \texttt{2} & Local Policy
			      & 0.40 & 9.44 & 7.60 & 6.31
		         && 0.26 & 7.09 & 5.63 & 4.84
		         && 0.40 & 8.04 & 6.36 & 5.73 \\
			\bottomrule
		\end{tabular}}
	\end{center}
\end{table}

\csubsection{Distribution of Navigation Errors}

In \tabref{tab:nav_errors} we present the navigation errors for both the Oracle Policy and Local Policy when used to perform the VLN-CE task. The presented short-range navigation errors were collected during the experiments reported in \tabref{tab:navigator}. We characterize the distribution of navigation errors by the percent of navigations that result in various thresholds of error. The Oracle Policy rarely produces a high navigation error -- just 0.53\% of navigations result in at least a 0.20m error in Val-Unseen. In the same setting, the Local Policy has a 0.20m failure rate of 8.04\%. This extends to even larger failure thresholds where the Local Policy fails to navigate to within 1.0m of the subgoal 5.73\% of the time in Val-Unseen. These failure rates provide context to \tabref{tab:navigator} which demonstrated a performance drop when navigating with the Local Policy.